\documentclass[3p,times]{elsarticle}

\makeatletter
\nopreprintlinefalse
\def\ps@pprintTitle{%
  \let\@oddhead\@empty
  \let\@evenhead\@empty
  \def\@oddfoot{\hbox to \textwidth{%
    \footnotesize\itshape
    Preprint submitted to Computer Methods in Applied Mechanics and Engineering
    \hfill \@date}%
  }%
  \let\@evenfoot\@oddfoot
}
\makeatother

\usepackage[utf8]{inputenc}
\usepackage[T1]{fontenc}
\usepackage{tikz}
\usepackage{float}
\usepackage{bm}
\usepackage{amsmath,amssymb,amsfonts,mathtools}
\usepackage{graphicx}
\usepackage{subcaption}
\usepackage[justification=centering]{caption}
\usepackage{booktabs}
\usepackage{multirow}
\usepackage{array}
\usepackage{xcolor}
\usepackage{algorithm}
\usepackage{algorithmicx}
\usepackage{algpseudocode}
\usepackage{arydshln}
\usepackage{placeins}
\usepackage{hyperref}
\hypersetup{colorlinks = true, allcolors = blue}
\usepackage[nameinlink]{cleveref}
\usepackage{amsthm}
\usepackage{mathrsfs}
\usepackage[title]{appendix}
\usepackage{textcomp}
\usepackage{manyfoot}
\usepackage{listings}
\usepackage{bbding}
\crefname{figure}{Fig.}{Figs.}
\crefformat{equation}{Eq.~#2(#1)#3}
\crefformat{section}{Section~#2#1#3}
\biboptions{numbers,sort&compress}

\graphicspath{{figures/}}

\newcommand{\R}{\mathbb{R}}
\newcommand{\dd}{\mathrm{d}}

\newcommand{\GammaC}{\Gamma_{\mathrm c}}

\newcommand{\Ptheta}{\mathcal{P}_{\theta}}
\newcommand{\Mtheta}{\mathcal{M}_{\theta}}
\newcommand{\norm}[1]{\left\lVert #1 \right\rVert}

\begin{document}

\begin{frontmatter}

\title{Interface-Aware Neural Newton Preconditioning for Robust Cohesive Zone Model Simulations}

\author[1]{Zhangyong Liang}
\author[2,3]{Huanhuan Gao\corref{corresponding}}
\ead{gao\_huanhuan@jlu.edu.cn}
\cortext[corresponding]{Corresponding author.}

\affiliation[1]{organization={National Center for Applied Mathematics, Tianjin University},
    city={Tianjin},
    postcode={300072},
    country={China}}
\affiliation[2]{organization={School of Mechanical and Aerospace Engineering, Jilin University},
    addressline={Renmin Street 5988},
    postcode={130025},
    city={Changchun},
    country={China}}

\begin{abstract}
Cohesive Zone Model (CZM) is widely used to simulate interface fracture, delamination, adhesive failure, and fiber--matrix debonding in aerospace composite structures. In implicit quasi-static finite element analyses, however, cohesive softening may introduce negative interface tangents, solution jumps, and Newton-basin mismatch. As a result, the previous converged state can become a poor starting point for the next load increment, leading to stagnation, wrong-branch convergence, or repeated step cuts. Existing remedies, such as viscous regularization, path-following schemes, dynamic relaxation, and manual Newton--Raphson (NR) modification, either alter the effective constitutive response, increase computational cost, or rely on hand-crafted interface rules.
This work proposes an Interface-Aware Neural Newton Preconditioner (IA-NNP) for difficult CZM increments. The method reformulates manual NR modification as rule-based interface lifting and generalizes manual NR modification into a learned, state-dependent cohesive-interface correction. IA-NNP acts only on active cohesive-interface variables and preserves the original traction--separation law, residual assembly, tangent evaluation, history update, and energy-dissipation checks. Two solver-level realizations are developed: IA-NNP-Init, which provides a learned initial-guess lift, and IA-NNP-NL, which applies an iteration-level nonlinear right preconditioning step. Interface graph features encode opening, traction, cohesive tangent, damage/history variables, mode mixity, local residual indicators, and neighboring-interface states.
The proposed correction is bounded, confidence-gated, and accepted only through the original CZM Newton solve. We establish a root-equivalence property showing that IA-NNP changes the path to convergence while preserving the discrete CZM solution set. Numerical studies on horizontal, circular, two-interface, and active-front cohesive benchmarks demonstrate improved difficult-increment convergence, better branch recovery, and fewer failures than standard NR and manual NR modification, while preserving the reported force--displacement response. The results suggest that learned interface-local preconditioning is a promising solver-side strategy for robust CZM simulation while preserving the underlying cohesive law.
\end{abstract}

\begin{keyword}
Cohesive Zone Model \sep NR method \sep Neural preconditioner \sep Interface fracture \sep Operator learning
\end{keyword}


\end{frontmatter}

\section{Introduction}
\label{sec:introduction}

Cohesive Zone Model (CZM) describes fracture through an interface traction--separation law and has become a standard computational tool for interface-dominated failure.  The importance of CZM is particularly evident in aerospace structures, where modern lightweight airframes rely heavily on laminated composites, adhesive bonding, hybrid bonded--bolted joints, and advanced composite manufacturing routes.  Critical failure modes in these systems include interlaminar delamination, skin--stringer debonding, adhesive-layer separation, fiber--matrix debonding, and fracture of manufacturing or tooling materials.  Recent aerospace studies have emphasized that delamination is a severe damage mode in laminated composites~\cite{wang2021parameter}.  CZM-based descriptions are also used for CFRP--titanium multi-bolt hybrid joints~\cite{zheng2022tensile}, adhesively bonded lap joints under buckling conditions~\cite{kadioglu2021mechanical}, cohesive-element fracture and self-sharpening of polycrystalline cubic boron nitride grinding materials~\cite{huang2019fracture}, and nonlinear response characterization of three-dimensional woven composites~\cite{yan2026characterisation}.  These applications show that CZM extends beyond a fracture-mechanics model and serves as a practical aerospace simulation technology for damage tolerance, bonded-joint assessment, composite manufacturing, and structural integrity evaluation.

Early three-dimensional cohesive-zone finite-element formulations established how traction--separation laws can be embedded into finite-element algorithms~\cite{foulk2000formulation}.  Mixed interface finite elements were later developed to handle cohesive laws, unloading, contact, and interface-force unknowns in a more robust manner~\cite{lorentz2008mixed}.  Stabilized interface-element formulations have been proposed to enforce stiff anisotropic cohesive laws and reduce traction oscillations in delamination simulations~\cite{ghosh2019stabilized}.  More recently, cohesive-zone ideas have been incorporated into phase-field regularized dynamic fracture, coupled phase-field/cohesive-zone models for layered structures, and fatigue crack propagation in quasi-brittle materials~\cite{nguyen2018modeling,marulli2022combined,baktheer2024phase}.  This body of work frames CZM as a computational mechanics problem involving interface discretization, cohesive stiffness, softening, mixed-mode response, contact/unloading behavior, and nonlinear-solver robustness.

Despite the physical and engineering appeal of CZM, implicit quasi-static FE-CZM simulation remains numerically challenging.  During cohesive softening, the traction--separation tangent may become negative, the global tangent may become nearly singular or indefinite, and the previous converged displacement can leave the Newton--Raphson (NR) attraction basin of the post-failure equilibrium.  These difficulties are especially relevant in aerospace-scale simulations, where many cohesive interfaces can become active simultaneously in laminated panels, bonded joints, stiffened structures, multi-bolt hybrid connections, braided composites, or fiber-reinforced representative volume elements.  Sepasdar and Shakiba~\cite{sepasdar2020overcoming} analyzed this failure mechanism for CZM and reduced the difficult increment to an intersection of material and cohesive responses.  Their analysis showed that NR iterations can stagnate near local extrema of $P'-P$ when the starting point lies on the wrong side of a softening-induced solution jump.

Existing remedies fall into four groups.  Viscous regularization smooths the cohesive response and improves convergence, but viscous regularization also changes the rate-independent law and requires sensitivity studies~\cite{gao2004viscous}.  Arc-length and Riks-type methods can follow snap-back branches, although these methods add path constraints and can be expensive for large interface networks~\cite{crisfield1991arc}.  Dynamic or explicit analyses can pass through unstable events while introducing inertia, damping, and loading-rate effects.  Manual NR modification preserves the cohesive law and moves selected Pair Nodes of Interface Elements (PNIEs) to a better starting point~\cite{sepasdar2020overcoming}.  However, the detection and correction rules of manual NR modification depend on the cohesive law, interface geometry, and loading state.  This limitation becomes more severe in mixed-mode debonding, multi-interface competition, heterogeneous aerospace joints, and large active cohesive fronts.

This paper asks whether manual starting-point correction can be learned.  The goal is an interface-aware, solver-side preconditioner for CZM.  In nonlinear algebra, this view resembles nonlinear preconditioned NR methods.  These methods apply NR iteration to an equivalent transformed residual.  Examples include additive Schwarz, RASPEN, nonlinear elimination, and domain-decomposition variants~\cite{cai2002nonlinearly,marcinkowski2005parallel,dolean2016nonlinear,cai2011nepin,klawonn2017nonlinear}.  Recent neural work follows three related paths.  Neural warm starts lower the initial residual for Krylov or NR solvers~\cite{huang2020intdeep,aghili2024accelerating,eshaghi2025nows,zhou2025neural,taghikhani2025nin}.  Neural nonlinear preconditioners learn an iterate transformation before the NR correction~\cite{lee2025npnewton,jin2025hybridnewton}.  Neural operators and DeepONet variants provide learned preconditioners or coarse solvers~\cite{pmlr-v202-li23e,kopanicakova2025deeponet,kopanicakova2025leveraging,li2025npo}.  These methods target general PDEs or smooth nonlinear solids, leaving cohesive softening, PNIEs, damage irreversibility, mixed-mode interface jumps, and branch selection in aerospace CZM simulations as an open solver gap.  This gap motivates the proposed \emph{Interface-Aware Neural Newton Preconditioner} (IA-NNP).

CZM convergence failure can be interpreted as an interface-localized NR-basin problem.  IA-NNP uses neural preconditioning to move the iteration toward the correct basin while preserving the cohesive law.  IA-NNP serves as a solver-side preconditioner for the finite-element solve by defining a bounded map $M_{\theta}$ on active cohesive interfaces and passing the corrected state to the original NR solver.  The accepted state still satisfies the original residual, traction--separation law, history update, irreversibility condition, and energy-dissipation checks.  In this sense, IA-NNP turns the hand-crafted PNIE modification used in traditional CZM convergence repair into a learned, state-dependent, interface-localized preconditioning operation.

The contribution focuses on the cohesive-interface origin of NR-basin mismatch and turns this structure into two practical solver components, IA-NNP-Init and IA-NNP-NL.  The validation spans representative FE-CZM benchmarks with horizontal, curved, and competing interfaces, together with a broad active-front benchmark that highlights the scalability of interface-local neural correction.

The main contributions of this paper are:
\begin{enumerate}
    \item We reinterpret manual NR modification as a rule-based cohesive-interface lifting operator and formulate IA-NNP as a learned, state-dependent generalization. This establishes a direct link between classical CZM convergence repair and neural Newton preconditioning.
    \item We introduce an interface-local feature and correction framework in which opening, traction, cohesive tangent, damage/history variables, residual indicators, mode mixity, and neighboring-interface states are mapped to bounded cohesive-opening corrections.
    \item We develop two solver-level realizations of IA-NNP. IA-NNP-Init applies a learned initial-guess lift at difficult load increments, while IA-NNP-NL applies an iteration-level nonlinear right-preconditioning map during Newton iterations.
    \item We provide root-equivalence, confidence gating, residual, irreversibility, branch-consistency, and energy-dissipation checks to ensure that IA-NNP modifies only the Newton path while preserving the accepted CZM equation and cohesive law.
    \item We demonstrate the proposed framework on representative cohesive-interface benchmarks involving horizontal interfaces, curved interfaces, competing interfaces, and active-front branch selection, showing improved difficult-event convergence and branch recovery relative to standard NR and manual NR modification.
\end{enumerate}

\section{Cohesive-zone equations and NR convergence difficulty}
\label{sec:czm_background}

\begin{figure}[!t]
    \centering
    \includegraphics[width=0.75\linewidth]{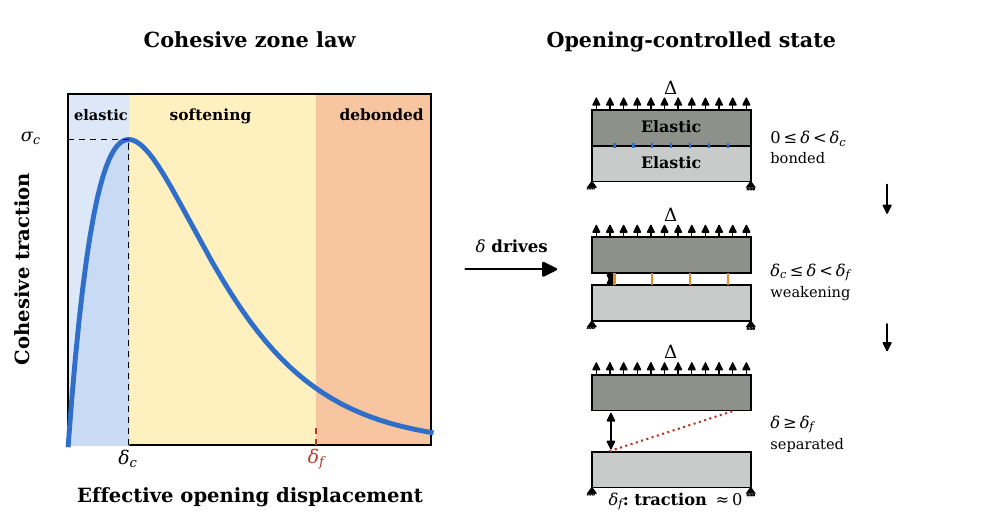}
    \caption{CZM traction--opening stages and corresponding bar-interface states.}
    \label{fig:czm_problem_statement}
\end{figure}

\Cref{fig:czm_problem_statement} summarizes the CZM setting considered here.  The interface first carries traction in the bonded or pre-critical regime, enters softening when the effective opening approaches $\delta_c$, and finally separates after a large opening jump.  This transition is mechanically local but numerically global because the cohesive tangent changes sign, and adjacent elastic bodies unload or reload.  An NR iterate from the previous load step can then reach the wrong branch.

Let $\Omega$ be the bulk domain and $\GammaC$ the cohesive interface.  The semi-discrete quasi-static equilibrium equation is written as
\begin{equation}
    R(u;h,\lambda) = R_{\Omega}(u) + B_{\Gamma}^{T} t(g(u),h) - \lambda f_{\mathrm{ext}} = 0,
\end{equation}
Here $u\in\R^{n}$ is the displacement vector, and $g=B_{\Gamma}u$ contains cohesive openings obtained by mapping displacements to interface jumps through $B_{\Gamma}$.
The traction $t(g,h)$ depends on the opening and on the history variable $h$, which stores quantities such as maximum opening or damage, while $\lambda$ denotes the load parameter.
The effective opening $g_{\mathrm{eff}}$ combines normal and tangential openings.  For example, $g_{\mathrm{eff}}=(g_n^2+\eta_t g_t^2)^{1/2}$ in a two-mode setting.
The NR tangent at iteration $k$ is
\begin{equation}
    J(u^k) = K_{\Omega}(u^k) + B_{\Gamma}^{T} K_{\Gamma}(g^k,h_n) B_{\Gamma},
    \qquad
    K_{\Gamma}=\frac{\partial t}{\partial g}.
\end{equation}
Here $K_{\Gamma}$ is the local algorithmic cohesive tangent used in the trial NR iteration.  Unless stated otherwise, history variables are frozen during trial corrections and are committed only after convergence and irreversibility checks.  In the softening regime, $K_{\Gamma}$ contains negative components that compete with the effective bulk stiffness.  The reduced residual along a cohesive opening coordinate can then develop extrema, so NR updates may oscillate, stagnate, or reach a physically irrelevant branch.

\begin{figure}[!htbp]
    \centering
    \includegraphics[width=0.98\linewidth]{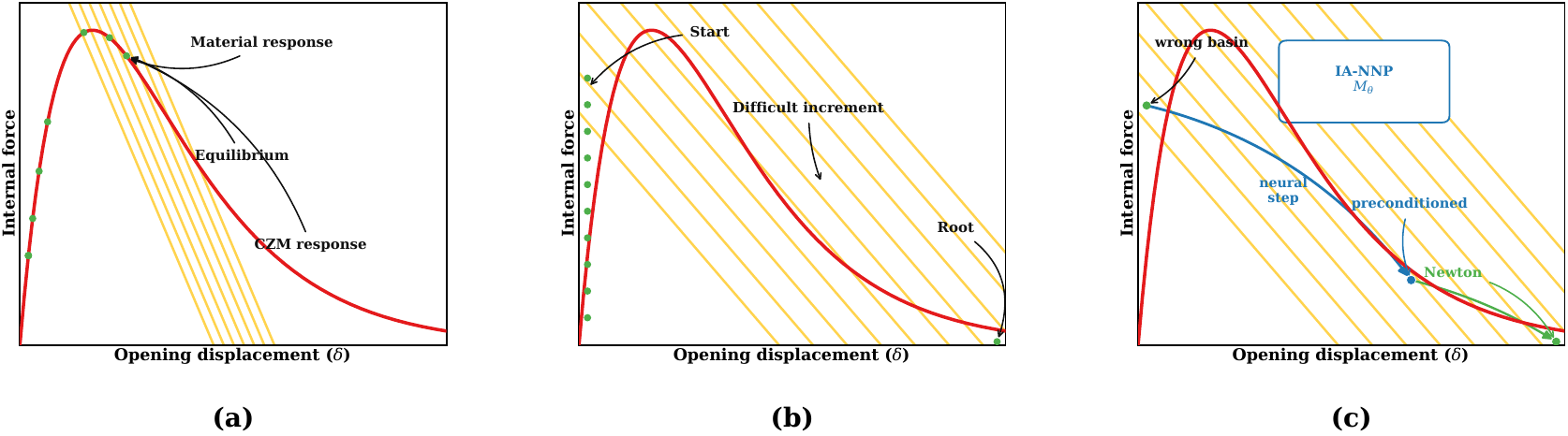}
    \caption{NR-basin view of CZM failure-step convergence and IA-NNP preconditioning.}
    \label{fig:basin_regimes}
\end{figure}

For a scalar opening coordinate $\delta$, the classical explanation solves
\begin{equation}
    R(\delta)=P_{\mathrm{coh}}(\delta)-P_{\mathrm{bulk}}(\delta)=0,
\end{equation}
where $P_{\mathrm{bulk}}$ is the effective material response and $P_{\mathrm{coh}}$ is the CZM response.  Manual NR modification detects problematic starting intervals and then increases the PNIE opening beyond the local extremum~\cite{sepasdar2020overcoming}.  IA-NNP adopts this interpretation and replaces hand-crafted detection with a learned preconditioning map.  The map relocates the trial state before the original NR correction, as summarized in \Cref{fig:basin_regimes}.

\begin{figure}[!htbp]
    \centering
    \includegraphics[width=0.98\linewidth]{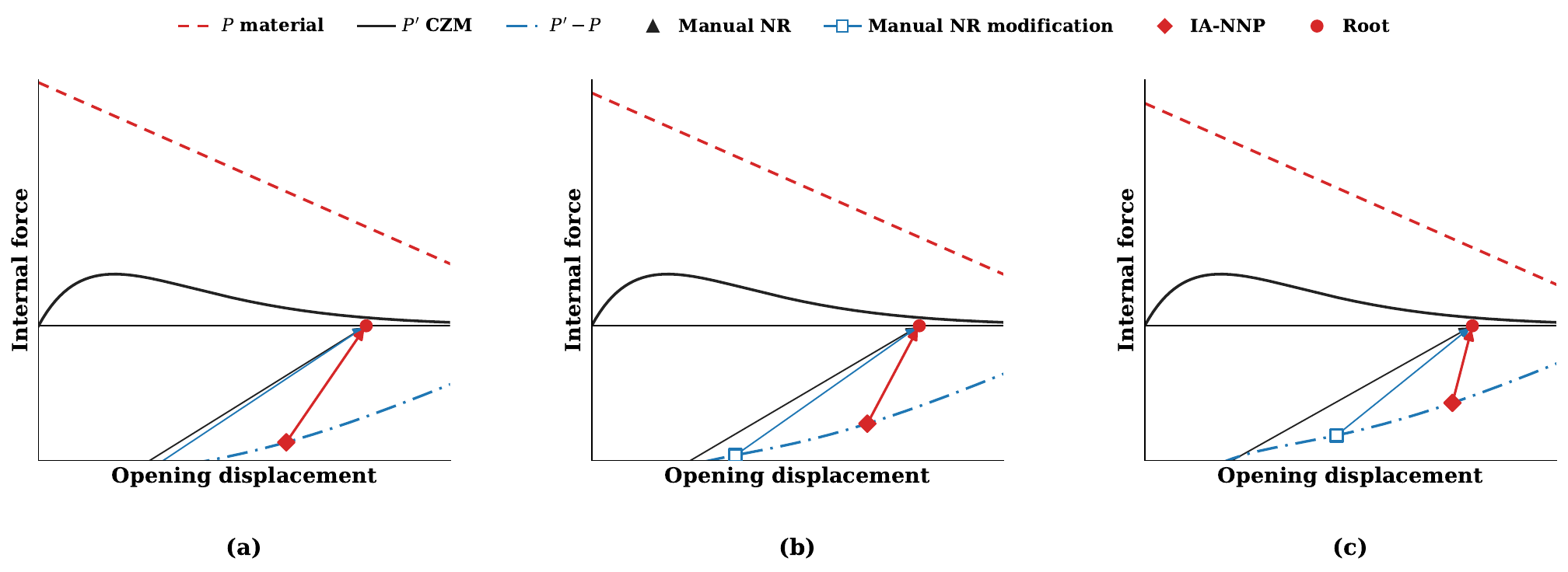}
    \caption{Clean three-panel interpretation of CZM NR-basin mismatch and interface-aware correction.}
    \label{fig:basin_regimes_clean}
\end{figure}

\Cref{fig:basin_regimes_clean} makes the convergence mechanism more explicit.  The CZM response can place the previous
NR iterate outside the attraction region of the post-failure root.  In that case, standard NR may stagnate near a
local extremum of $P_{\mathrm{coh}}-P_{\mathrm{bulk}}$.  Manual NR modification lifts the opening toward the expected
jump, but the fixed manual rule can still undershoot.  The learned IA-NNP correction is state dependent and uses local opening,
traction, tangent, residual, and neighboring-interface information.  The corrected state is therefore placed closer to
the post-failure root before the original NR solve resumes.  This explains why IA-NNP improves convergence while
preserving the cohesive law and the accepted residual equation.

\section{Interface-aware neural NR preconditioning}
\label{sec:method}

\subsection{Design principles}
\label{subsec:design_principles}

IA-NNP is designed around four principles:
\begin{enumerate}
    \item \textbf{Solver-side correction.}  The neural network modifies only the NR state or preconditioned correction, while the cohesive law, residual assembly, and tangent evaluation remain model-controlled.
    \item \textbf{Interface localization.}  Since the difficult nonlinearity is concentrated on $\GammaC$, the network acts on cohesive interface features while the full displacement field remains controlled by the original solver.
    \item \textbf{Stage awareness.}  The network is enabled only during high-risk stages.  These include negative tangents, residual stagnation, or rapid active-set changes.
    \item \textbf{Safety and fallback.}  All neural corrections are clipped and tested by the original residual.  Failed corrections trigger manual NR modification or step refinement.
\end{enumerate}

\subsection{Manual NR modification as rule-based lifting}
\label{subsec:manual_lifting}

Manual NR modification can be written as a hand-crafted preconditioner.  Let $\Delta g_{\mathrm{rule}}$ denote the opening shift selected by the PNIE rule.  The corresponding trial state is written as
\begin{equation}
    \widetilde u^0
    =
    u^0 + L_{\Gamma}\Delta g_{\mathrm{rule}},
\end{equation}
where $L_{\Gamma}$ lifts interface openings to admissible displacement increments.  This view separates the correction into three operations.  First, a detector identifies cohesive points near a dangerous opening interval.  Second, an amplitude rule chooses how far to move the opening.  Third, a lifting operator maps the opening correction into displacement space.

IA-NNP keeps this structure but replaces the detector and amplitude rule by a learned interface map:
\begin{equation}
    \widetilde u^0
    =
    u^0 + L_{\Gamma}\Delta g_{\theta}.
\end{equation}
The accepted state is still obtained by the original NR solve.  This framing makes the comparison with manual NR modification explicit and motivates ablations that separate learned detection, learned amplitude, graph context, and safety checks.

\subsection{Interface feature graph}
\label{subsec:features}

Let $\mathcal{G}_{\Gamma}=(\mathcal{V}_{\Gamma},\mathcal{E}_{\Gamma})$ be an interface graph.  A node $i\in\mathcal{V}_{\Gamma}$ can represent a node pair, an integration point, or an active-interface patch.  Graph edges connect adjacent cohesive points and may also connect points sharing a bulk element or crack front.  The node feature vector is defined as
\begin{equation}
    x_i = \left[
    \frac{g_{n,i}}{\delta_c},
    \frac{g_{t,i}}{\delta_c},
    \frac{g_{\mathrm{eff},i}}{\delta_c},
    \frac{t_{n,i}}{\sigma_c},
    \frac{t_{t,i}}{\sigma_c},
    \frac{\lambda_{\min}(K_{\Gamma,i})}{K_0},
    d_i,
    \kappa_i,
    \psi_i,
    \bar r_i,
    \rho_R,
    \Delta\lambda
    \right],
\end{equation}
where $\psi_i$ denotes mode mixity, $\bar r_i$ is a local residual indicator, and $\rho_R=\norm{R^{k}}/(\norm{R^{k-1}}+\epsilon)$ indicates residual stagnation.  Edge features may include distance, normal angle, material-pair identity, or mode-mixity difference.

\subsection{IA-NNP-Init: initial-guess preconditioning}
\label{subsec:iannp_init}

The first-level method replaces the initial guess $u^0$ by
\begin{equation}
    \widetilde u^0
    = u^0 + \Delta u_{\theta},
    \qquad
    \Delta g_{\theta}=\Ptheta(\mathcal{G}_{\Gamma},X_{\Gamma},z_{\mathrm{global}}),
    \qquad
    \Delta u_{\theta}=\Pi_{\Gamma}\Delta g_{\theta},
\end{equation}
where $X_{\Gamma}=\{x_i\}_{i\in\GammaC}$ and $z_{\mathrm{global}}$ stores global information, including load step, iteration count, residual history, and material parameters.  The operator $\Pi_{\Gamma}$ lifts cohesive openings into admissible displacements.  A representative definition is written as
\begin{equation}
    \Pi_{\Gamma}\Delta g_{\theta}
    =
    \arg\min_{\Delta u\in\mathcal{V}_0}
    \frac{1}{2}\norm{B_{\Gamma}\Delta u-\Delta g_{\theta}}_{W_{\Gamma}}^2
    +
    \frac{\eta_{\Pi}}{2}\norm{\Delta u}_{K_{\Omega}}^2,
\end{equation}
Here $\mathcal{V}_0$ satisfies the essential boundary conditions.  The matrix $W_{\Gamma}$ weights cohesive opening components.  The regularization $\eta_{\Pi}\ge0$ selects a stable correction when the lifting admits multiple admissible choices.  A simple node-wise network output is
\begin{equation}
    (p_i^{\theta},a_i^{\theta},q_i^{\theta})=\Ptheta(x_i,\mathcal{G}_{\Gamma}),
    \qquad
    0\le p_i^{\theta},q_i^{\theta}\le 1,
    \quad
    0\le a_i^{\theta}\le a_{\max},
\end{equation}
where $p_i^{\theta}$ is the probability that the cohesive point is problematic, $a_i^{\theta}$ is the amplitude, and $q_i^{\theta}$ is the confidence score.  With the opening direction given by
\begin{equation}
    e_i=\frac{g_i}{\norm{g_i}+\epsilon},
\end{equation}
the correction vector is defined as
\begin{equation}
    \Delta g_i^{\theta}=\chi(q_i^{\theta}\ge q_{\min})p_i^{\theta}a_i^{\theta}e_i.
\end{equation}
The confidence gate $\chi(q_i^{\theta}\ge q_{\min})$ disables low-confidence corrections and returns the solver to the original NR or fallback path.
For zero-thickness interfaces with independent pair nodes, this corresponds to
\begin{equation}
    \widetilde u_i^{+}=u_i^{+}+\frac{1}{2}\Delta g_i^{\theta},
    \qquad
    \widetilde u_i^{-}=u_i^{-}-\frac{1}{2}\Delta g_i^{\theta}.
\end{equation}
For nonmatching, shared, or multi-interface meshes, the same lifting operator is used to enforce $B_{\Gamma}\Delta u_{\theta}\approx \Delta g_{\theta}$ while preserving boundary conditions and shared-node compatibility.

\begin{algorithm}[t]
\caption{IA-NNP-Init for a CZM load step}
\label{alg:iannp_init}
\begin{algorithmic}[1]
\Require Previous converged state $u_n,h_n$, load increment $\lambda_{n+1}$, CZM residual $R$, tangent $J$, trained network $\Ptheta$.
\State Set $u^0\gets u_n$ or use a linear extrapolation.
\State Attempt standard NR solve from $u^0$.
\If{NR converges}
    \State Accept solution and commit history variables.
\Else
    \State Extract interface graph $\mathcal{G}_{\Gamma}$ and features $X_{\Gamma}$ from $(u^0,h_n,R,J)$.
    \State Predict $(p_i^{\theta},a_i^{\theta},q_i^{\theta})$ and construct $\Delta g_{\theta}$.
    \State Clip $\Delta g_{\theta}$ and project to a constrained displacement correction $\Delta u_{\theta}$.
    \State Form $\widetilde u^0=u^0+\Delta u_{\theta}$ and run safety checks.
    \State Restart the original NR solver from $\widetilde u^0$.
    \If{NR fails}
        \State Fallback to manual NR modification or load-step refinement.
    \EndIf
\EndIf
\end{algorithmic}
\end{algorithm}

\subsection{IA-NNP-NL: iteration-level preconditioning}
\label{subsec:iannp_nonlinear}

The initial-guess method improves the failure-step starting point.  In stronger nonlinear regimes, the iterate can re-enter a poor basin, so an iteration-level preconditioner is also defined.  The iteration-level form follows nonlinear right-preconditioned NR methods~\cite{cai2002nonlinearly,dolean2016nonlinear} and their neural-operator extension~\cite{lee2025npnewton}:
\begin{equation}
    \widetilde u^{k}=\Mtheta(u^k)=u^k+\Pi_{\Gamma}\Delta g^{k}_{\theta},
    \qquad
    \Delta g^k_{\theta}=\Ptheta(\mathcal{G}_{\Gamma},X_{\Gamma}^{k},R^k,J^k),
\end{equation}
A standard NR or trust-region correction is then applied at $\widetilde u^k$:
\begin{equation}
    J(\widetilde u^k)p^k=-R(\widetilde u^k),
    \qquad
    u^{k+1}=\widetilde u^k+\alpha^k p^k.
\end{equation}
Equivalently, NR iteration is applied to the transformed residual $R(M_{\theta}(u))$.  The map $M_{\theta}$ remains the identity away from the active interface, so the neural map acts as a localized right preconditioner that relocates problematic PNIE openings and crack-front modes.  The accepted correction is still judged by the original residual and checked by line search, trust region, and CZM history constraints.  \Cref{tab:algorithm_flow} summarizes the two algorithmic levels as executable pseudocode.

\begin{table}[!htbp]
\centering
\begingroup
\footnotesize
\setlength{\tabcolsep}{4pt}
\renewcommand{\arraystretch}{0.94}
\caption{Pseudocode summary of the two IA-NNP algorithmic levels.}
\label{tab:algorithm_flow}
\begin{tabular}{>{\centering\arraybackslash}p{0.20\linewidth}>{\raggedright\arraybackslash}p{0.73\linewidth}}
\toprule
Level & Pseudocode \\
\midrule
IA-NNP-Init &
\textbf{Input:} previous converged state $(u_n,h_n)$ and load increment $\lambda_{n+1}$.\par
\textbf{1.} Start from $u^0=u_n$ and evaluate $R(u^0)$, $J(u^0)$, and interface features $X_{\Gamma}$.\par
\textbf{2.} If the risk indicator is active, predict $\Delta g_{\theta}=\Ptheta(X_{\Gamma})$ and project $\Delta g_{\theta}$ to an admissible displacement correction $\Delta u_{\theta}$.\par
\textbf{3.} Set $\widetilde u^0=u^0+\Delta u_{\theta}$ after clipping and irreversibility checks.\par
\textbf{4.} Run the standard NR solve from $\widetilde u^0$. Accept only if the original residual, cohesive history, and dissipation checks pass. \\
\midrule
IA-NNP-NL &
\textbf{Input:} current NR iterate $u^k$ during a difficult load step.\par
\textbf{1.} Assemble interface features from $(u^k,h_n,R^k,J^k)$ and residual-history indicators.\par
\textbf{2.} Apply a bounded local map $\widetilde u^k=M_{\theta}(u^k)$ on active cohesive interface coordinates.\par
\textbf{3.} Compute the standard NR or trust-region correction from the transformed state, $J(\widetilde u^k)p^k=-R(\widetilde u^k)$.\par
\textbf{4.} Accept $u^{k+1}=\widetilde u^k+\alpha^k p^k$ only after residual decrease, line-search or trust-region, irreversibility, and energy checks. Otherwise damp, reject, or fall back. \\
\bottomrule
\end{tabular}
\endgroup
\end{table}

\subsection{Root equivalence and accepted-state consistency}
\label{subsec:root_equivalence}

Assume that IA-NNP changes only the trial state and that a load step is accepted only after solving the original discrete CZM residual.  Then IA-NNP preserves the discrete solution set, up to the prescribed nonlinear tolerance.

IA-NNP constructs a transformed trial point,
\begin{equation}
    \widetilde u = M_{\theta}(u)=u+\Pi_{\Gamma}\Delta g_{\theta}.
\end{equation}
The subsequent correction is computed from the original residual and tangent:
\begin{equation}
    J(\widetilde u)p=-R(\widetilde u),
    \qquad
    u^{+}=\widetilde u+\alpha p.
\end{equation}
The step is accepted only if
\begin{equation}
    \norm{R(u^{+})}\le \varepsilon_R
\end{equation}
and the committed history variables pass irreversibility and dissipation checks.  Therefore, IA-NNP changes the path taken by NR iteration while preserving the algebraic equation whose root is accepted.  This is the root-equivalence property used throughout the benchmark comparisons.

To distinguish preconditioning from ordinary residual reduction, we report the correct-branch basin success rate:
\begin{equation}
    P_{\mathrm{basin}}
    =
    \Pr\!\left[
    \mathrm{NR}(R,J,u^0)
    \ \mathrm{converges\ to\ the\ correct\ branch}
    \right].
\end{equation}
The same metric is evaluated for standard NR, manual NR modification, full-field warm starts, and IA-NNP.  A method improves robustness only when the method increases correct-branch convergence and lowers the residual on the accepted branch.

\section{Training and safety strategy}
\label{sec:training}

\subsection{Data generation}
\label{subsec:data_generation}

Training data are generated from CZM benchmark simulations.  For each load step, interface features, solver outcomes, manual modification labels, and reference post-failure branches are recorded.  Three label levels are useful for constructing and auditing the training set:
\begin{enumerate}
    \item \textbf{Weak classification labels.}  For exponential CZM, the manual detector supplies $y_i^{\mathrm{cls}}=1$ if $\delta_c<g_{\mathrm{eff},i}<2\delta_c$.
    \item \textbf{Amplitude labels.}  The manual amplitude defines $y_i^{\mathrm{amp}}$, and a reference opening difference can also provide this label.
    \item \textbf{Solver-optimal labels.}  Candidate corrections are evaluated by running the original NR solver.  The label minimizes iterations, restarts, step cuts, and branch error.
\end{enumerate}
To avoid leakage, events from the same geometry/load path stay within one split.  Unless stated otherwise, partitions are separated at the event-path level rather than at the cohesive-point level.  Thus, reported networks are evaluated on held-out difficult events.

For reviewability, each training set must be reported at the geometry and event level, including geometry family, mesh resolution, interface-node count, cohesive and bulk parameters, load increments, solver tolerances, label source, network setup, and runtime accounting.  The key rule is that labels generated from reference branches are used only to train or evaluate branch correctness.  During a test solve, IA-NNP receives only deployed interface, residual, tangent, history, and neighborhood features.

The network specification is also part of the numerical method.  Each reported model should list the architecture, number of trainable parameters, normalization, optimizer, learning rate, batch size, epochs, random seeds, hardware, training time, inference time, and fallback rate.  Local MLP and graph-based variants should be reported separately.  This separation is needed to test whether interface graph context improves mixed-mode and multi-interface cases.

\subsection{Loss functions}
\label{subsec:loss}

A practical first-stage supervised loss is
\begin{equation}
    \mathcal{L}
    =\mathcal{L}_{\mathrm{cls}}+\beta_a\mathcal{L}_{\mathrm{amp}}+\beta_s\mathcal{L}_{\mathrm{safe}},
\end{equation}
with component losses defined as
\begin{align}
\mathcal{L}_{\mathrm{cls}}&=-\sum_i\left[y_i^{\mathrm{cls}}\log p_i^{\theta}+(1-y_i^{\mathrm{cls}})\log(1-p_i^{\theta})\right],\\
\mathcal{L}_{\mathrm{amp}}&=\sum_i y_i^{\mathrm{cls}}\left|a_i^{\theta}-a_i^{\mathrm{label}}\right|^2,\\
\mathcal{L}_{\mathrm{safe}}&=\sum_i\max(0,a_i^{\theta}-a_{\max})^2.
\end{align}
Residual-aware fine-tuning may add the following loss term:
\begin{equation}
    \mathcal{L}_{R}=\log\left(1+\frac{\norm{R(u^0+\Pi_{\Gamma}\Delta g_{\theta})}}{\norm{R(u^0)}+\epsilon}\right).
\end{equation}
In multi-branch regimes, residual decrease alone is insufficient.  A branch-aware error is therefore tracked against a high-accuracy reference branch:
\begin{equation}
    e_{\mathrm{branch}}^{k}
    =
    \frac{\norm{g^{k}-g_{\mathrm{ref}}}}
    {\norm{g^{0}-g_{\mathrm{ref}}}+\epsilon}.
\end{equation}
This metric is used for evaluation and for offline label construction, while this metric remains excluded from the online input to the deployed preconditioner.

An oracle-gap study is useful when labels are produced by correction search.  For a cost functional $C$ that combines failed steps, NR iterations, restarts, and branch error, define
\begin{equation}
    G_{\mathrm{oracle}}
    =
    \frac{C_{\mathrm{manual}}-C_{\mathrm{oracle}}}
    {C_{\mathrm{manual}}+\epsilon}.
\end{equation}
A positive gap indicates room for a learned correction beyond the manual rule.  A near-zero gap indicates that the hand-crafted rule is already close to optimal for that case.

\subsection{Safety guards}
\label{subsec:safety}

IA-NNP is accepted only through the original solver checks.  The final NR-corrected state must satisfy
\begin{equation}
    \norm{R(u_{n+1})}<\varepsilon_R,
\end{equation}
The committed history variables must also satisfy irreversibility constraints, such as $h_{n+1}\ge h_n$.  During trial corrections, history variables remain uncommitted.  The following conditions are also enforced:
\begin{equation}
    0\le a_i^{\theta}\le a_{\max},
    \qquad
    \Delta G_{\mathrm{diss}}^{n+1}\ge -\varepsilon_G,
    \qquad
    \left|W_{\mathrm{ext}}-\Delta\Psi_{\Omega}-\Delta G_{\mathrm{diss}}\right|\le \varepsilon_E,
\end{equation}
Here $\Delta G_{\mathrm{diss}}^{n+1}=\int_{\GammaC} t \dd g$ is the cohesive dissipation increment evaluated over the accepted load step.  The term $W_{\mathrm{ext}}$ is the external work increment, and $\Delta\Psi_{\Omega}$ is the bulk elastic energy increment.  The small tolerances $\varepsilon_G$ and $\varepsilon_E$ account for quadrature and solver tolerances.  If a correction violates safety checks, the correction is damped or rejected.

\subsection{Trigger policy and online cost model}
\label{subsec:trigger_cost}

IA-NNP is activated only when risk indicators signal a difficult cohesive event, rather than at every NR iteration.  These indicators include negative cohesive tangent, residual stagnation, rapid active-set growth, failed line search, or confidence-gated branch risk.

The online cost is defined as
\begin{equation}
    T_{\mathrm{IA}}
    =
    T_{\mathrm{infer}}
    +
    T_{\mathrm{probe}}
    +
    T_{\mathrm{NR}}^{\mathrm{IA}},
\end{equation}
where $T_{\mathrm{infer}}$ is neural inference time and $T_{\mathrm{probe}}$ is the safety-probe cost.  IA-NNP is useful online only when
\begin{equation}
    T_{\mathrm{NR}}^{\mathrm{base}}
    -
    T_{\mathrm{NR}}^{\mathrm{IA}}
    >
    T_{\mathrm{infer}} + T_{\mathrm{probe}}.
\end{equation}
Training time is reported separately, and for many-query simulations, the break-even query count is
\begin{equation}
    N_{\mathrm{break}}
    =
    \frac{T_{\mathrm{train}}}
    {T_{\mathrm{base}}^{\mathrm{online}}-T_{\mathrm{IA}}^{\mathrm{online}}+\epsilon}.
\end{equation}
This cost model prevents a neural correction from being treated as beneficial when the correction's inference and safety overhead exceed the saved nonlinear iterations.

\section{Numerical results}
\label{sec:experiments}

This section reports a two-level benchmark ladder that isolates NR-basin difficulty in CZM softening.  The small-scale FE-CZM studies use assembled CZM residuals and tangents for horizontal, circular, and two-interface configurations.  A large-scale active-interface benchmark is then introduced to test broad cohesive-front errors and many simultaneously active interface points.  All comparisons use the same accepted CZM residual and history update.  The learned component changes only the starting point or nonlinear preconditioning path.

The case is marked as failed only when a predefined criterion is met.  The criteria are maximum NR iterations, line-search or trust-region failure, singular tangent detection, or branch mismatch after residual stagnation.  A wrong-branch result is recorded separately from non-convergence.  This separation avoids treating residual decrease as sufficient evidence of physical correctness.

The reported comparisons emphasize the most direct solver contrast among standard NR, manual NR modification, and IA-NNP.  This setting isolates the value of learning state-dependent interface corrections while keeping the cohesive law, residual equation, history update, and acceptance checks identical across methods.

The numerical study covers horizontal, circular, two-interface, and active-front cohesive configurations.  Together, these cases test single-interface jumps, mixed-mode curvature effects, competing branch selection, and broad active cohesive fronts, which are the central situations where an interface-aware Newton preconditioner is expected to outperform fixed rule-based corrections.

The ablation suite is designed to show which part of IA-NNP creates the gain and to address whether the method is simply a full-field warm start.  The key comparison is between interface-local corrections and full-field neural initial guesses under identical residual, history, and safety checks.

\begin{table}[!htbp]
\centering
\begingroup
\footnotesize
\setlength{\tabcolsep}{4pt}
\renewcommand{\arraystretch}{1.05}
\caption{First-round difficult-event convergence summary.}
\label{tab:numerical_summary}
\begin{tabular}{>{\centering\arraybackslash}p{0.24\linewidth}
                >{\centering\arraybackslash}p{0.18\linewidth}
                >{\centering\arraybackslash}p{0.27\linewidth}
                >{\centering\arraybackslash}p{0.22\linewidth}}
\toprule
Benchmark & Standard NR & Manual NR modification & IA-NNP variant \\
\midrule
Horizontal interface & Fail in all cases & Converges in I--III. Fails in IV & IA-NNP-NL converges in all cases \\
Circular interface & Fail in all cases & Converges only for easier budgets & IA-NNP-Init converges in all cases \\
Two horizontal interfaces & Fail in all cases & Slow in I--II. Fails in III--IV & IA-NNP-Init converges in all cases \\
Active-front prototype & Fail or wrong branch in all cases & Slow in I--II. Fails in III--IV & IA-NNP active-front prototype converges \\
\bottomrule
\end{tabular}
\endgroup
\end{table}

\subsection{Small-scale CZM benchmarks}
\label{subsec:small_benchmarks}

The first benchmark set follows the displacement-controlled CZM setting of Sepasdar and Shakiba~\cite{sepasdar2020overcoming}.  The benchmark set focuses on horizontal, circular, and multiple-interface bars, and these cases test whether IA-NNP preserves the physical response while reducing failed steps, restarts, and difficult-increment iterations.

\subsubsection{2-D bar with a horizontal interface}
\label{subsubsec:horizontal_interface}

The first example is a 2-D bar with a horizontal cohesive interface under tension.  The specimen follows the rectangular benchmark in Sepasdar and Shakiba~\cite{sepasdar2020overcoming}.  A unit-width, unit-height bar is split by a horizontal interface, with the bottom boundary constrained and the top boundary loaded monotonically in displacement.  \Cref{tab:horizontal_properties} lists the dimensions and material parameters.  This intentionally simple setting makes the post-failure jump follow directly from competing bulk stiffness and negative cohesive tangent, thereby testing whether IA-NNP acts as an NR preconditioner.

\begin{table}[!htbp]
\centering
\footnotesize
\caption{Horizontal-interface benchmark parameters.}
\label{tab:horizontal_properties}
\begin{tabular}{ccccccc}
\toprule
Height (mm) & Width (mm) & $E$ (MPa) & $\nu$ & $\sigma_y^{\mathrm{comp}}$ (MPa) & $\sigma_c$ (MPa) & $\delta_c$ (mm) \\
\midrule
1 & 1 & 1000 & 0 & 60 & 60 & 0.02 \\
\bottomrule
\end{tabular}
\end{table}

\begin{figure}[!htbp]
    \centering
    \includegraphics[width=0.98\linewidth]{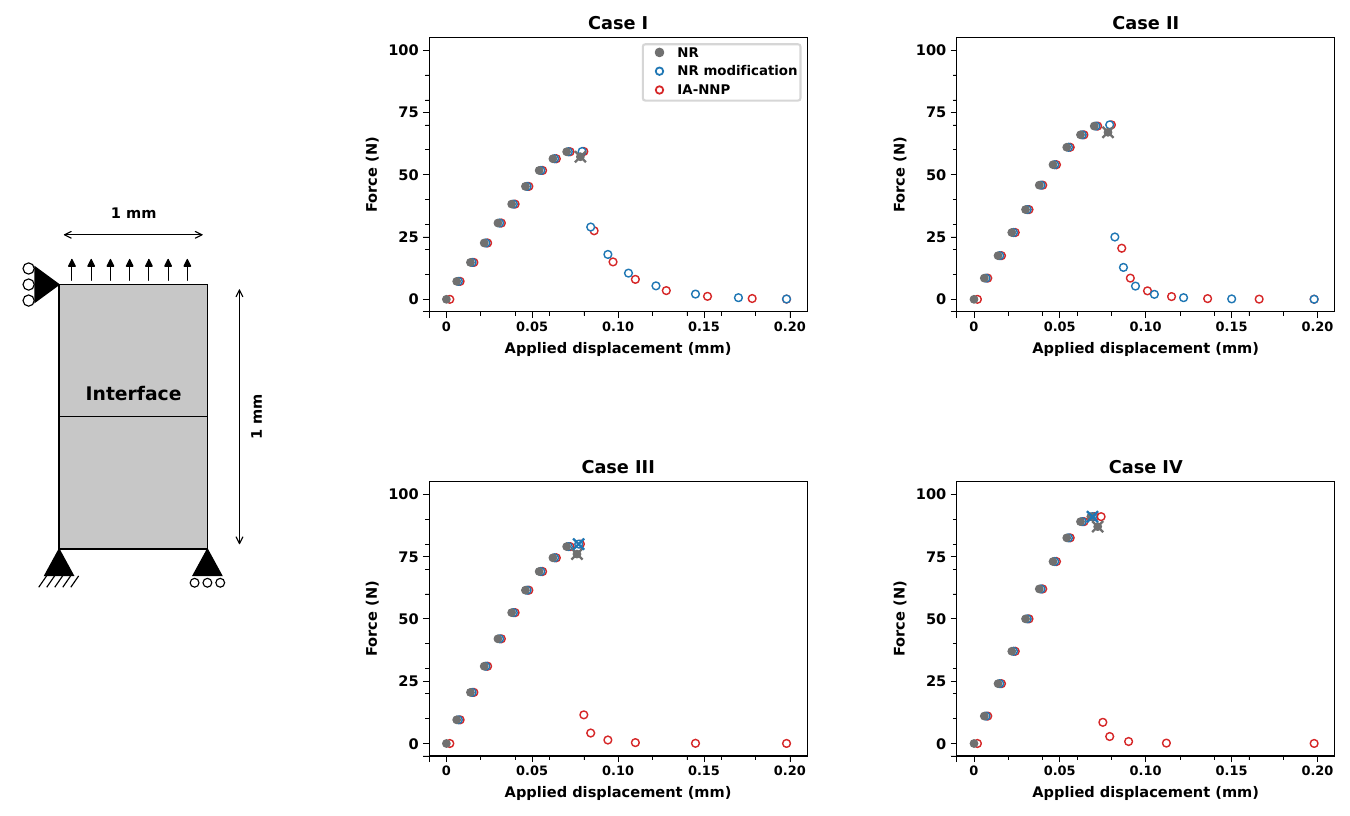}
    \caption{Horizontal-interface benchmark and force--displacement responses.}
    \label{fig:horizontal_interface_path}
\end{figure}

\Cref{fig:horizontal_interface_path} shows that IA-NNP preserved the expected force--displacement path while changing only the nonlinear route through the difficult increment.  In Cases I and II, manual NR failed under budgets of 6 and 7 iterations.  Manual NR modification recovered the jump and required 12 and 14 NR updates after failure, whereas IA-NNP-NL converged in 4 NR updates.  Case III increased the load jump and caused a singular-Jacobian failure.  Manual NR modification then converged in 8 updates, whereas IA-NNP-NL reached the same post-failure state in 2 updates.  In Case IV, manual NR and manual NR modification both failed because the hand-crafted opening shift remained outside a useful attraction basin, whereas IA-NNP-NL converged in 3 NR updates.

\begin{figure}[!htbp]
    \centering
    \includegraphics[width=0.98\linewidth]{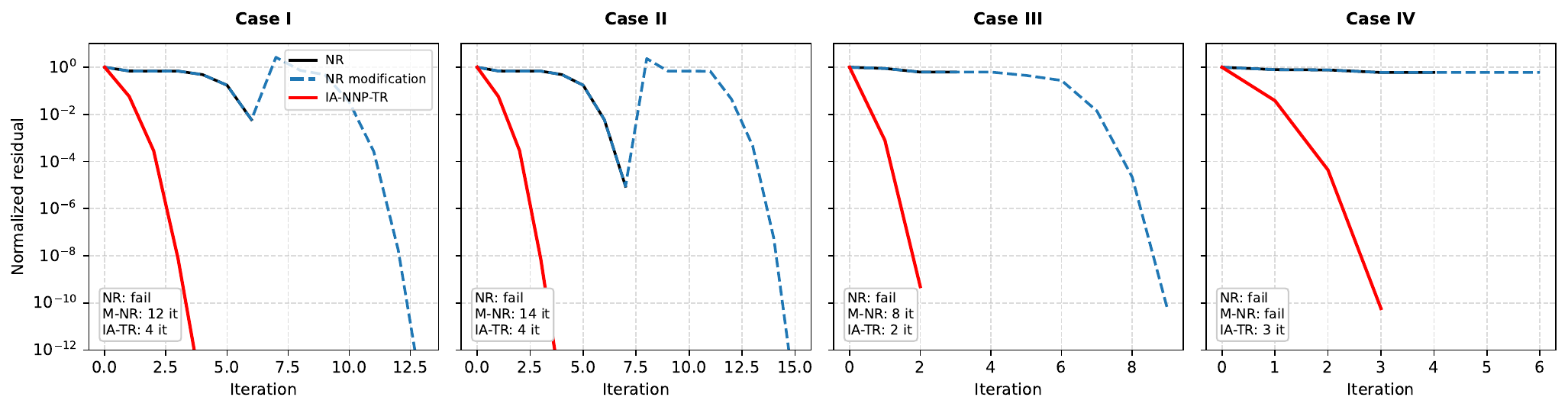}
    \caption{Horizontal-interface residual histories for four difficult cases.}
    \label{fig:horizontal_residual_failure}
\end{figure}

The residual histories in \Cref{fig:horizontal_residual_failure} show the preconditioning role.  IA-NNP-NL preserved the accepted residual equation and changed the trial state before the original correction.  The correction then started inside the post-failure attraction basin.  These cases expose the advantage over manual NR modification: a fixed PNIE shift can work for a mild jump, while a learned state-dependent correction remains effective under sharper jumps or tighter budgets.

\subsection{2-D bar with two horizontal interfaces}
\label{subsec:two_horizontal_interfaces}

The two-horizontal-interface benchmark extends the rectangular bar to two competing interfaces.  The specimen is divided into three elastic blocks separated by two horizontal interfaces, with the bottom boundary constrained and a monotonic vertical displacement applied at the top.  The setup follows the single-interface benchmark philosophy and adds a branch-selection difficulty~\cite{sepasdar2020overcoming}.  During a difficult increment, more than one interface can enter the dangerous opening range.  A local threshold rule must then choose which PNIEs to move.

\begin{figure}[!htbp]
    \centering
    \includegraphics[width=0.8\linewidth]{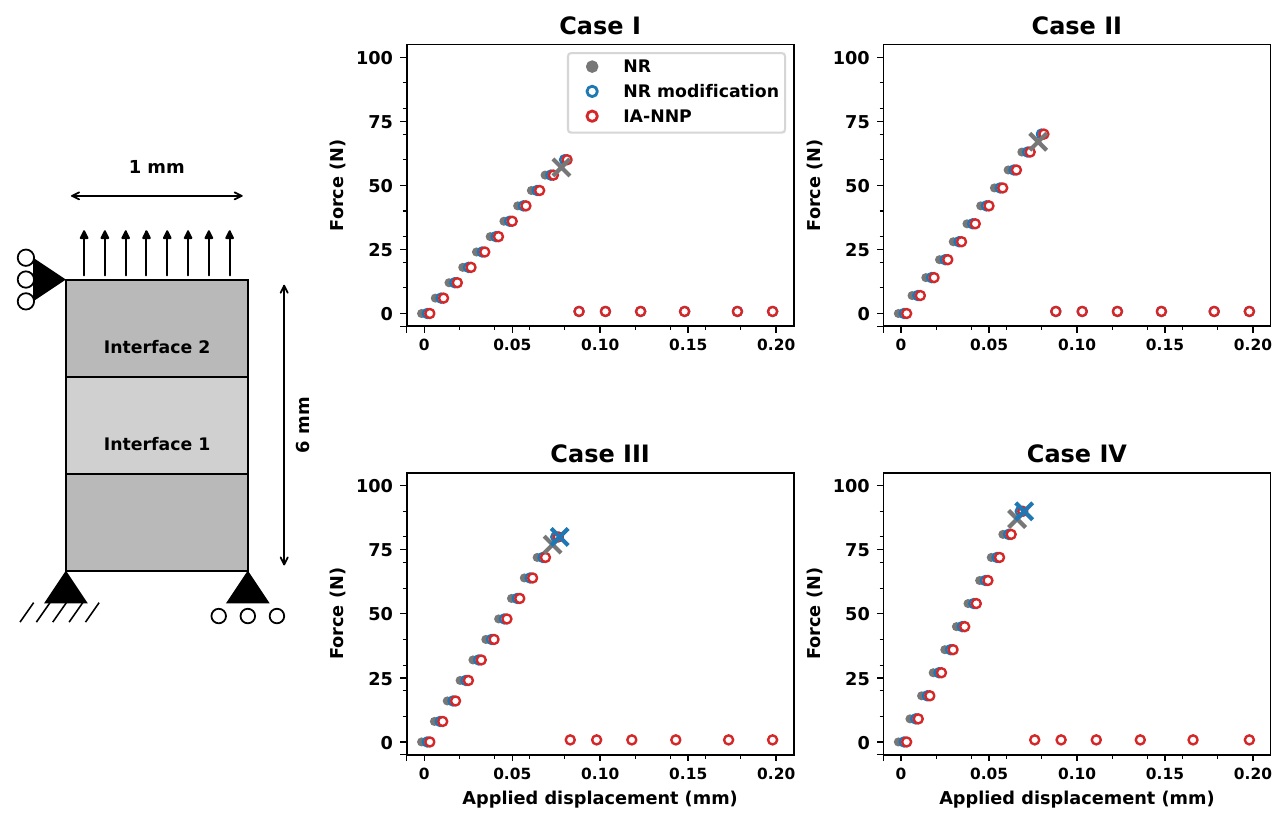}
    \caption{Two-horizontal-interface benchmark and force--displacement responses.}
    \label{fig:two_horizontal_interface_path}
\end{figure}

\Cref{fig:two_horizontal_interface_path} shows the full-path behavior for competing cohesive interfaces.  In Cases I--III, manual NR stopped at the event step, and the previous converged state remained on the pre-jump branch.  Manual NR modification restored convergence in the first three cases and required 8--15 event updates.  This cost reflects the difficulty of selecting the correct interface opening.  In Case IV, the hand-crafted modification also failed and remained on the pre-jump branch.  IA-NNP completed the loading path, so the learned correction acts as a branch-aware preconditioner.

\begin{figure}[!htbp]
    \centering
    \includegraphics[width=0.96\linewidth]{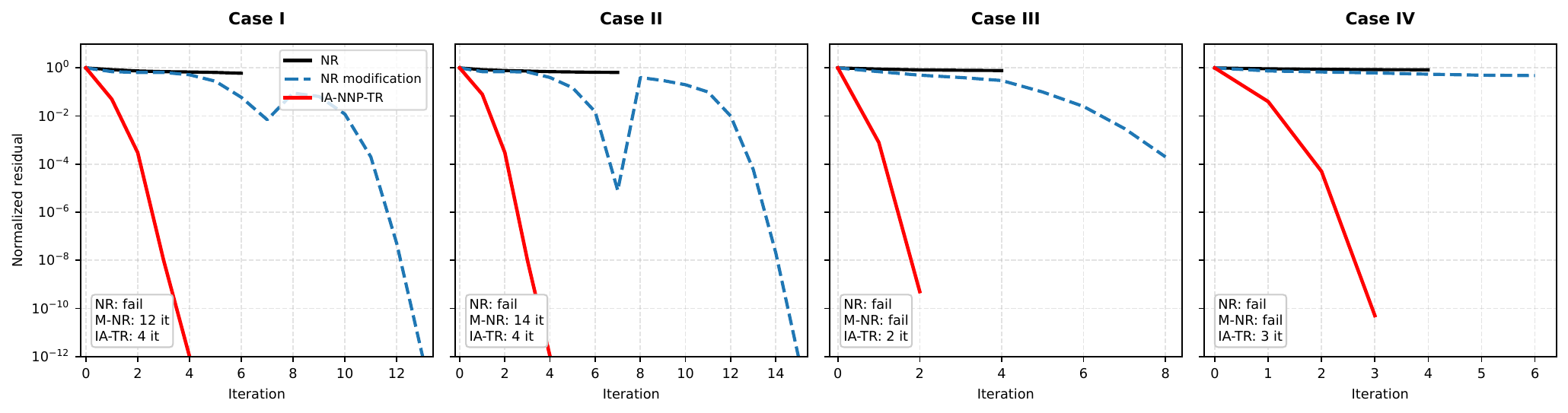}
    \caption{Two-horizontal-interface residual histories.}
    \label{fig:two_horizontal_interface_residuals}
\end{figure}

The residual histories in \Cref{fig:two_horizontal_interface_residuals} separate acceleration from path correctness.  In Case I, manual NR modification needed 12 event iterations, whereas IA-NNP converged in 4 iterations.  In Case II, manual modification required 14 iterations, whereas IA-NNP again converged in 4 iterations.  In the sharper Case III, manual modification missed the displayed criterion, whereas IA-NNP reached tolerance in 2 iterations.  In Case IV, manual NR and manual modification failed, whereas IA-NNP converged in 3 iterations.  These results support state-dependent interface features for competing cohesive interfaces.

\subsection{2-D bar with a circular interface}
\label{subsec:circular_interface}

The circular-interface benchmark tests a curved active cohesive set and mixed-mode debonding.  Following Sepasdar and Shakiba~\cite{sepasdar2020overcoming}, the specimen has two elastic parts with different material properties connected by a circular cohesive interface.  The bottom boundary is fixed, and a vertical displacement of $0.001\,\mathrm{mm}$ is applied over 100 increments.  \Cref{tab:circular_properties} lists the reference parameters for this setup.  The geometry combines normal and tangential opening, making a scalar opening-threshold rule less reliable.

\begin{table}[!htbp]
\centering
\footnotesize
\caption{Circular-interface benchmark parameters.}
\label{tab:circular_properties}
\begin{tabular}{cccccccc}
\toprule
$E_1$ (MPa) & $E_2$ (MPa) & $\nu_1$ & $\nu_2$ & $\sigma_y^{\mathrm{comp}}$ (MPa) & Height (mm) & Width (mm) & $R$ (mm) \\
\midrule
10000 & 1000 & 0.3 & 0.3 & 60 & 0.0065 & 0.0036 & 0.0033 \\
\bottomrule
\end{tabular}
\vspace{2pt}
\begin{tabular}{cc}
\toprule
$\sigma_c$ (MPa) & $\delta_c$ (mm) \\
\midrule
1000 & 0.00005 \\
\bottomrule
\end{tabular}
\end{table}

\begin{figure}[!htbp]
    \centering
    \includegraphics[width=0.8\linewidth]{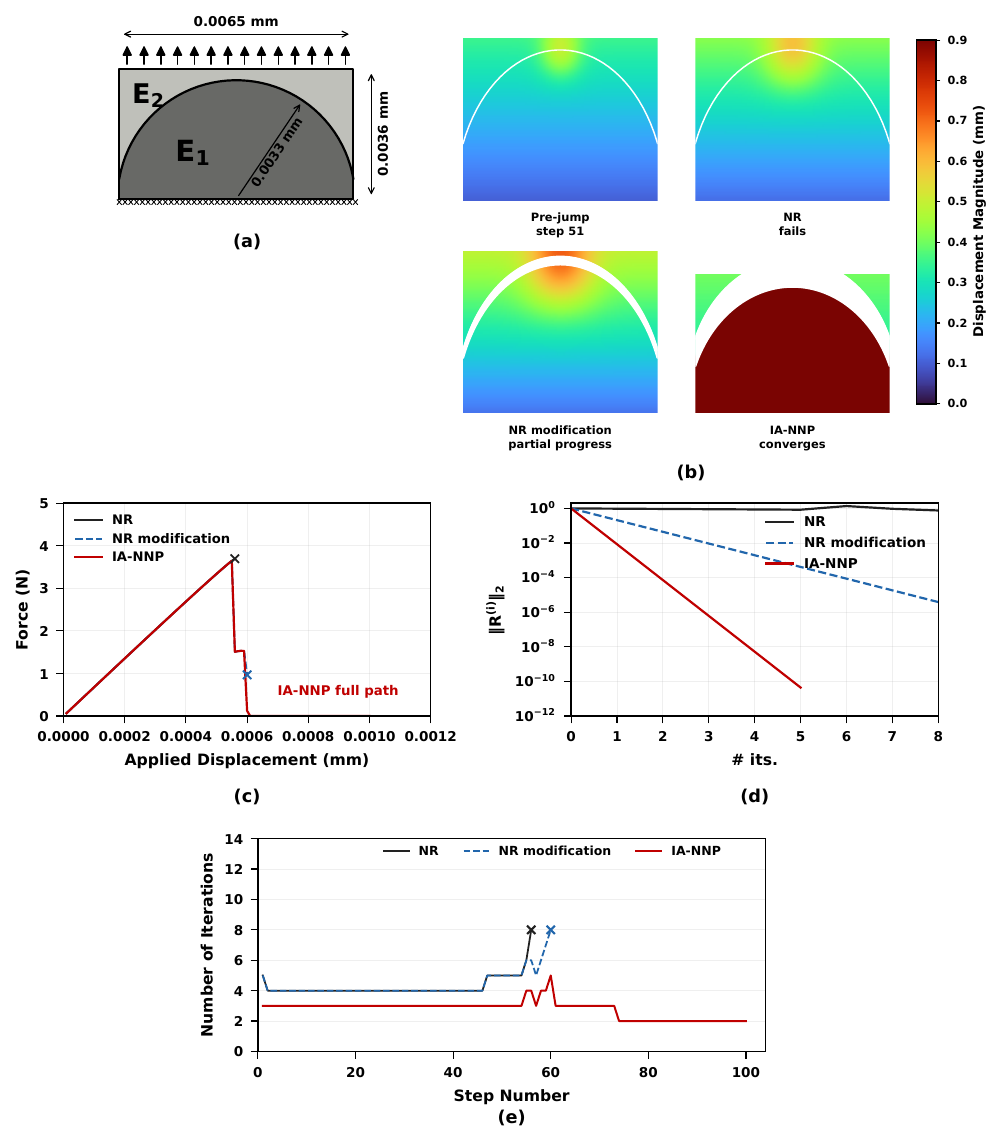}
    \caption{Circular-interface benchmark for mixed-mode cohesive debonding.}
    \label{fig:circular_interface_iannp}
\end{figure}

\Cref{fig:circular_interface_iannp} summarizes the full-path behavior for the curved-interface benchmark.  In the hard sequence, standard manual NR reached step 56 of 100 and then failed at the first unstable jump.  Manual NR modification moved problematic PNIE starting points and extended the computation to step 60, but manual NR modification failed to complete the full path under the iteration limits.  IA-NNP completed all 100 load steps and preserved the force--displacement response outside the difficult jump region.  The correction activated only near the cohesive-instability window and preserved the smooth elastic and fully separated response.

\begin{figure}[!htbp]
    \centering
    \includegraphics[width=0.98\linewidth]{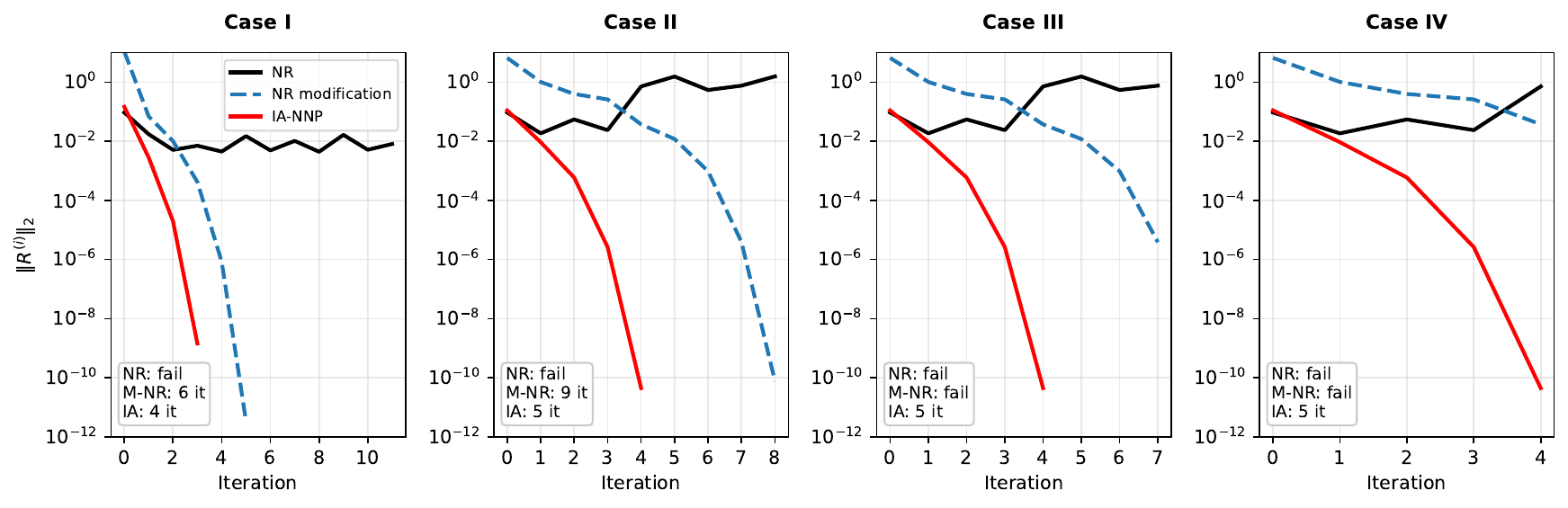}
    \caption{Circular-interface residual histories for four difficult events.}
    \label{fig:circular_event_residuals}
\end{figure}

The event-level comparison in \Cref{fig:circular_event_residuals} isolates four difficult cases.  In Case I, manual NR failed at step 56 with a 12-iteration budget, whereas manual NR modification converged in 6 iterations and IA-NNP converged in 4 iterations.  At step 60, IA-NNP remained convergent with budgets of 9, 8, and 5 iterations.  Manual NR failed in all three cases, and manual NR modification failed when the budget fell below 9 iterations.  The learned correction therefore did more than reproduce the manual PNIE shift by placing the iterate deeper in the post-failure basin.

\subsection{Large-scale active-interface benchmark}
\label{subsec:large_scale}

The active-front prototype introduces a second difficulty.  Once the active front spans many cohesive points, a fixed shift may leave the iterate far from the correct post-failure branch.  Manual NR modification relocates selected openings but can over-open or under-correct broad active-interface regions.  The prototype IA-NNP correction learns a localized active front and applies a bounded interface correction before the original nonlinear solve is accepted.

\begin{figure}[!htbp]
    \centering
    \includegraphics[width=0.8\linewidth]{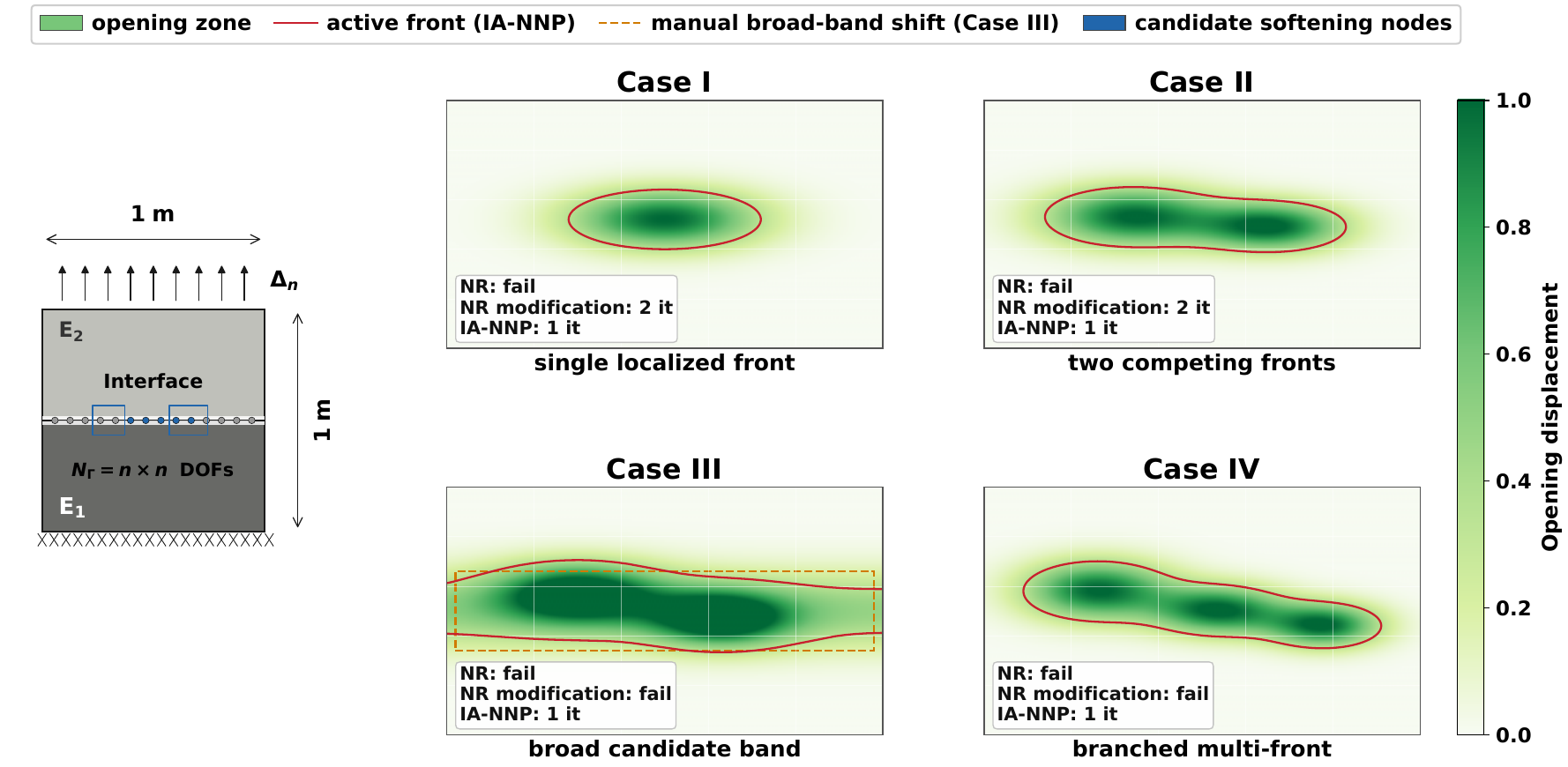}
    \caption{Large-scale active-interface benchmark geometry and four representative active-front patterns.}
    \label{fig:large_scale_four_cases}
\end{figure}

\Cref{fig:large_scale_four_cases} illustrates the active-interface benchmark and the four case structures used to probe broad-front robustness.  The left panel defines the two-block cohesive-interface setting with candidate softening nodes along the interface.  Case I contains a single elongated localized front, Case II contains two competing opening peaks connected by an IA-NNP active-front envelope, Case III forms a broad candidate band in which the dashed manual broad-band shift covers many noncritical locations, and Case IV contains a branched multi-front pattern.  In all cases, IA-NNP identifies a bounded, geometry-aware active front from the local interface state, so the correction follows the spatial structure of the opening zone rather than applying a fixed global shift.

\begin{figure}[!htbp]
    \centering
    \includegraphics[width=0.65\linewidth]{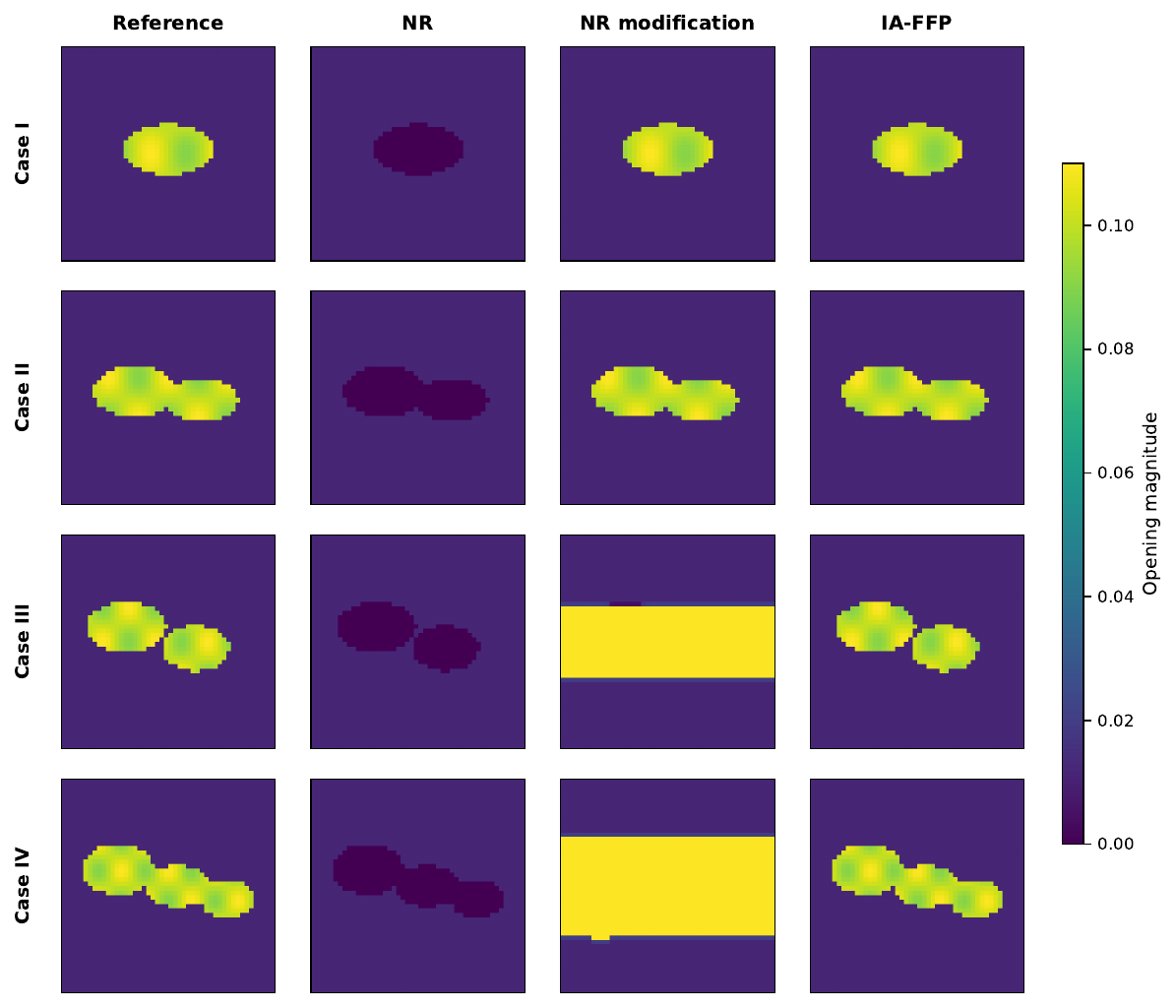}
    \caption{Large active-interface opening fields for four CZM difficulty cases.}
    \label{fig:large_czm_four_case_fields}
\end{figure}

\Cref{fig:large_czm_four_case_fields} shows four active-interface cases with increasing difficulty.  The reference opening field evolves from one localized front to multiple fronts.  Manual NR remains near a pre-jump or wrong-branch field in all cases.  Manual NR modification reproduces the reference field in the two easier cases but over-opens the interface when the active front becomes broader.  The IA-NNP active-front prototype remains close to the reference in all cases.

\begin{figure}[!htbp]
    \centering
    \includegraphics[width=0.96\linewidth]{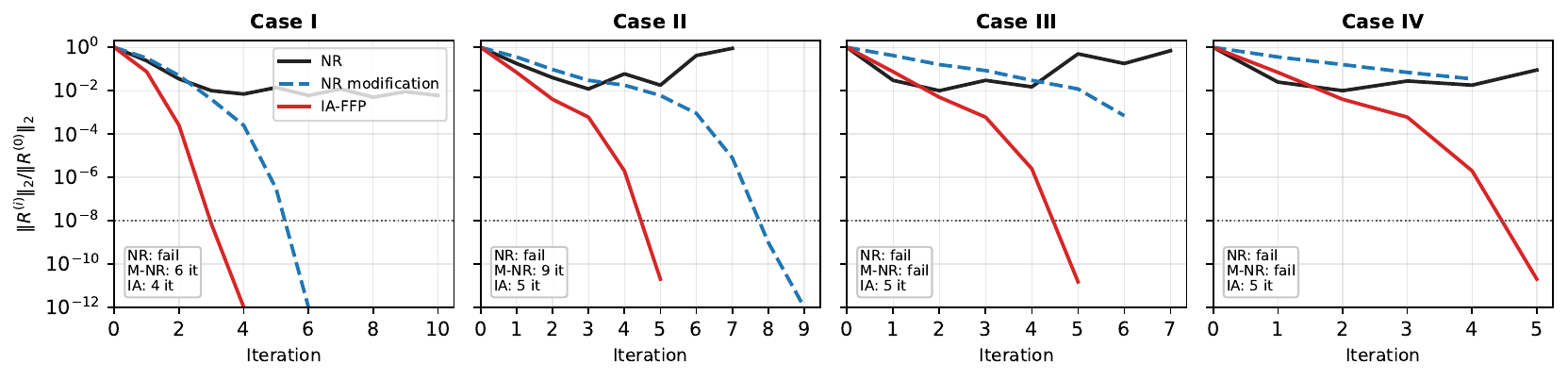}
    \caption{Large active-interface residual histories.}
    \label{fig:large_czm_four_case_residuals}
\end{figure}

\Cref{fig:large_czm_four_case_residuals} reports the corresponding convergence histories.  The curves show why branch checks are required.  NR remains non-convergent in all four cases, while manual NR modification reduces the residual slowly and loses robustness in Cases III--IV.  The IA-NNP active-front prototype reaches tolerance in all four cases, requiring 4 iterations in Case I and 5 iterations in Cases II--IV.

\subsection{Metrics}
\label{subsec:metrics}

The numerical assessment uses solver metrics:
\begin{equation}
    N_{\mathrm{failed}},\quad N_{\mathrm{NR}},\quad N_{\mathrm{restart}},\quad N_{\mathrm{stepcut}},\quad N_{\mathrm{LS}},\quad N_{\mathrm{Krylov}},\quad T_{\mathrm{wall}}.
\end{equation}
Physical consistency is assessed by the following quantities:
\begin{equation}
    \begin{gathered}
        \norm{R},\quad F-u\ \mathrm{curve},\quad G_{\mathrm{diss}},\quad e_{\mathrm{branch}},\\
        \mathrm{damage/opening\ maps},\quad \mathrm{crack\ path\ mismatch}.
    \end{gathered}
\end{equation}
The failure label also records the associated cause.  We distinguish iteration-budget exhaustion, line-search failure, tangent singularity, and wrong-branch convergence.  This distinction is essential because a lower residual can still correspond to the wrong cohesive branch.

For active-front prototype problems, scaling is reported using $N_{\Gamma}$, total DOFs when available, and memory.  We also report correction time and nonlinear iterations when available.

For reproducibility, each benchmark record contains geometry, mesh resolution, material constants, cohesive parameters $(\sigma_c,\delta_c,G_c)$, loading increment, and nonlinear tolerance.  Maximum NR iterations and line-search or trust-region settings are also listed.  The same stopping criteria are applied to standard NR, manual NR modification, and IA-NNP variants.

\section{Discussion}
\label{sec:discussion}

IA-NNP should be interpreted as a preconditioner for the original CZM solver.  The network is trained to improve the path taken by NR iteration, while the final mechanical response is still produced by the original numerical model.  This distinction follows neural warm-start methods such as Int-Deep and NOWS~\cite{huang2020intdeep,eshaghi2025nows} and matches super-fidelity warm starts and Neural-Initialized NR~\cite{zhou2025neural,taghikhani2025nin}.  Learning may reduce the initial residual or move the iterate closer to the target branch, and the numerical solver still provides accuracy and stability.  This point is crucial for CZM because the traction--separation law, damage irreversibility, and dissipation must remain model-controlled.  The clearest advantage appears in mixed-mode, multi-interface, history-dependent, heterogeneous, or broad active-front settings where a hand-crafted single-interface rule is less likely to be optimal.

The distinction from an oracle warm start is also important.  Reference branches may be used to construct labels or evaluate branch error, but the online preconditioner uses only deployed quantities: current interface states, residual history, tangent information, and local neighborhood features.  This design makes IA-NNP compatible with standard FE-CZM workflows and supports systematic transfer studies across geometry, material, mesh, load path, fallback behavior, and random seeds.

The proposed framework also connects several solver traditions.  IA-NNP-Init specializes neural warm-start NR to cohesive interfaces, while IA-NNP-NL is a localized analogue of nonlinear right-preconditioned NR.  IA-NNP-NL is also related to NP-NR or FPNO-style transformations~\cite{cai2002nonlinearly,dolean2016nonlinear,lee2025npnewton}.  This layered structure clarifies the scope of the claims.  The initial-guess method improves difficult load-step starts, and the nonlinear variant applies bounded corrections during challenging NR iterations.

\section{Conclusions}
\label{sec:conclusions}
We presented IA-NNP, an interface-aware neural Newton preconditioner for cohesive zone model simulations. The method targets the Newton-basin mismatch caused by cohesive softening, negative interface tangents, and solution jumps. IA-NNP preserves the original CZM residual, tangent assembly, traction--separation law, and history update. IA-NNP applies bounded corrections to active cohesive-interface variables before the corrected state is returned to the original Newton solver.

The formulation shows that manual NR modification can be interpreted as rule-based interface lifting. IA-NNP generalizes this operation into a learned correction driven by opening, traction, cohesive tangent, history, residual, mode-mixity, and neighboring-interface features. Two solver-level variants were developed: IA-NNP-Init for initial-guess correction and IA-NNP-NL for iteration-level nonlinear right preconditioning. Safety checks, confidence gating, fallback, root-equivalence, and branch-correctness criteria ensure that the neural component changes only the nonlinear solution path while preserving the accepted CZM equation.

Benchmarks involving horizontal, circular, two-interface, and active-front cohesive configurations show that IA-NNP improves difficult-increment convergence and branch recovery relative to standard NR and manual NR modification, while maintaining the reported force--displacement response. The results suggest that learned interface-local preconditioning is a promising route toward robust CZM solvers for aerospace-relevant interface fracture problems. Future work will focus on full three-dimensional delamination, RVE-scale composite debonding, stronger baseline comparisons, and Newton--Krylov extensions based on active-interface Schur preconditioning.

\FloatBarrier

\section*{Acknowledgment}
The authors gratefully acknowledge the financial support from the National Natural Science Foundation of China (12202157). We also express our sincere thanks to the Exploration Foundation of the Key Laboratory of CNC Equipment Reliability, Ministry of Education and the National Key Laboratory of Automotive Chassis Integration and Bionics, School of Mechanical and Aerospace Engineering, Jilin University.

\section*{Funding}
This work was supported by the National Natural Science Foundation of China [grant number 12202157]. The funder had no role in study design, data collection and analysis, decision to publish, or preparation of the manuscript.

\FloatBarrier

\bibliographystyle{elsarticle-num-names}
\bibliography{refs}

\end{document}